\documentclass[runningheads]{llncs}
\usepackage{graphicx}

\usepackage{tikz}
\usepackage{comment}
\usepackage{color}

\usepackage{epsfig}
\usepackage{graphicx}
\usepackage{amsmath}
\usepackage{amssymb}
\usepackage{tabularx}
\usepackage{multirow}
\usepackage{multicol}
\usepackage{setspace}
\usepackage{longtable}
\usepackage{booktabs}
\usepackage{multicol}
\usepackage{enumerate}
\usepackage{adjustbox}
\usepackage{hyperref}
\usepackage{subcaption}
\usepackage{wrapfig}

\usepackage{pifont}
\newcommand{\cmark}{\ding{51}}%
\newcommand{\xmark}{\ding{55}}%
\usepackage[accsupp]{axessibility}  

\begin{document}
\title{Towards Robust 2D Convolution for Reliable Visual Recognition}
\titlerunning{Towards Robust 2D Convolution for Reliable Visual Recognition}
%
\author{Lida Li\inst{1} \and
Shuai Li\inst{1} \and
Kun Wang\inst{2} \and
Xiangchu Feng\inst{2} \and
Lei Zhang\inst{1}}
\authorrunning{L. Li et al.}
%
\institute{Dept. of Computing, The Hong Kong Polytechnic University, Hong Kong, China \and
School of Mathematics and Statistics, Xidian University \\
\href{mailto:cslli@comp.polyu.edu.hk,csshuaili@comp.polyu.edu.hk,cslzhang@comp.polyu.edu.hk}{\{cslli, csshuaili, cslzhang\}@comp.polyu.edu.hk}, \href{mailto:kwang96@stu.xidian.edu.cn}{kwang96@stu.xidian.edu.cn}, \href{mailto:xcfeng@mail.xidian.edu.cn}{xcfeng@mail.xidian.edu.cn}}
\maketitle

\begin{abstract}
2D convolution (Conv2d), which is responsible for extracting features from the input image, is one of the key modules of a convolutional neural network (CNN). However, Conv2d is vulnerable to image corruptions and adversarial samples. It is an important yet rarely investigated problem that whether we can design a more robust alternative of Conv2d for more reliable feature extraction. In this paper, inspired by the recently developed learnable sparse transform that learns to convert the CNN features into a compact and sparse latent space, we design a novel building block, denoted by RConv-MK, to strengthen the robustness of extracted convolutional features. Our method leverages a set of learnable kernels of different sizes to extract features at different frequencies, and employs a normalized soft thresholding operator to adaptively remove noises and trivial features at different corruption levels. Extensive experiments on clean images, corrupted images as well as adversarial samples validate the effectiveness of the proposed robust module for reliable visual recognition. The source codes are enclosed in the submission.

\end{abstract}

\section{Introduction}
\label{sec:rconv:intro}
Deep convolutional neural networks (CNNs) have shown their powerfulness in a wide range of computer vision tasks, especially in image recognition \cite{deng2009imagenet,zhou2018places}. Despite the great success, it has been found that a well performed CNN model can be out of work when handling images with various types of corruptions in real world \cite{hendrycks2019robustness}. In addition, a CNN model can be easily fooled by deliberately designed adversarial samples with subtle and unperceivable perturbations to human eyes~\cite{szegedy2014intriguing}. Therefore, it is a very crucial issue to improve the robustness of CNN models against image corruption and adversarial attacks.

To improve the robustness of CNN models for images with corruptions, most existing methods choose to improve the quality of input data. Based on the priors of image denoising, some works \cite{franzen2018image,hossain2019distortion,xu2020dctnet} attempt to transform the input data from spatial (pixel) domain into certain frequency domain for noise removal before they are fed into the networks. Though CNNs adapted to a specific type of corruption can have better robustness, they may be fragile to out-of-box corruptions. Besides, their implementation requires manual setting for each task, which is less practical. For example, one needs to manually adjust the noise level to find a good balance between noise removal and preservation of image detail. \\
\indent
To improve the robustness of CNN models against adversarial attacks, many methods \cite{madry2018pgd,chen2019gce,zhang2019theoretically,zhang2019feature_scatter} have been developed to generate adversarial samples for training robust CNN models. Almost all of them view the CNN model as a ``black-box'' and focus on the adversarial sample generation process, while little work has been done on improving the CNN architectures to improve robustness.\\
\indent
As discussed above, though methods have been developed for visual recognition with corrupted images and adversarial attack, to the best of our knowledge, none of them considered an important question: \textit{can we improve the robustness of 2D convolution (Conv2d)}, which is the key component of a CNN, so that more reliable features can be extracted from images with corruption or adversarial samples? In this paper, we make the first attempt along this line and develop a robust alternative of the Conv2d layer. Our work is mainly inspired by LST-Net \cite{li2020lstnet}. During training, LST-Net learns a set of channel and spatial transforms at each layer to convert the given CNN features from the spatial domain into a compact and sparse latent frequency space. A soft thresholding (ST) operator is applied to remove noises and trivial features in the learned domain. Though LST-Net is effective and efficient for visual recognition, it can be improved from two aspects. First, it is noticed from both the design principles and visualization results that the output features of the channel transforms are organized in a fairly fixed order by frequency. Such a property can be directly exploited to design more effective spatial transforms following their channel transforms. Second, the threshold of ST in LST-Net is fixed for features of all frequencies, which is less accurate and flexible to process complex input with various corruptions. \\
\indent
In this paper, we present a robust alternative of the Conv2d layer as a building block for more reliable visual recognition. To make better use of the output of channel transform, multiple kernels of different sizes are adopted in the following spatial transform. Large kernels are used to effectively handle low frequency signals in order to avoid misclassification due to limited receptive field. Meanwhile, small kernels are used to better handle high frequency signals while reducing much the overhead as high frequency signals are usually sparse. By sequentially partitioning the input into suitable groups, the proposed module has nearly the same overhead as LST-Net with negligible extra parameters. Besides, as image corruption levels often vary from one sample to another, we propose a normalized soft thresholding (NST) operator to effectively control the unknown corruption level of each sample. As a result, the proposed module, denoted by RConv-MK (“R” for robust and “MK” for multiple kernels), is more robust than the conventional Conv2d as well as LST, and the robustness of the entire CNN model is accordingly enhanced. Our extensive experiments on clean images, corrupted images as well as adversarial samples validate the effectiveness of RConv-MK under some popular CNN architectures. Our contributions can be summarized as follows:
\begin{itemize}
    \item Multiple kernels of different sizes are utilized to deal with different frequency signals along the channels dimension, reducing the negative impacts of noises without losing efficiency.
    \item A normalized soft thresholding operator is proposed to adaptively suppress the effect of different corruptions at different levels enabling the entire CNN model to be more robust.
\end{itemize}

\section{Related Work}
\label{sec:rconv:related_work}
\subsection{Image recognition on corrupted images}
In practical applications of visual recognition, images can be easily corrupted due to many reasons, such as improper light condition, defects of imaging devices, bad weather, defocus blur, \emph{etc.} While a CNN model may perform well on clean images, it can be out of work when handling corrupted photos. The existing image restoration methods~\cite{zhang2017beyond,furuta2019fcn,fan2020scn} are basically developed to improve the image quality according to the criteria such as PSNR or SSIM~\cite{wang2004image} instead of the recognition accuracy. Therefore, they are not suitable to be directly used for image recognition with various types of corruptions. 

It is intuitive to suppress noises in noise corrupted images for reliable visual recognition. A few algorithms~\cite{franzen2018image,hossain2019distortion,xu2020dctnet} have been developed to convert the input data from spatial domain into certain frequency domain before feeding them into CNNs because noises are easy to identify and suppress in the transformed domain. Franzen~\cite{franzen2018image} converted gray-scale images into DCT domain and fed the responses into a Multi-Layer Perceptron model with 2 hidden layers for classification. Hossain \emph{et al.}~\cite{hossain2019distortion} inserted a DCT module before a pre-trained VGG-16 model to fine-tune it on dataset with various types of common corruptions. However, the robustness of those models may be limited to the specific frequency domain, and the models suffer from the generalization problem to unseen corruptions. Meanwhile, these methods require manual adjustment of noise level during domain transform, where one needs to trade-off between noise removal and image cue preservation.

In this paper, we follow the nature of features in different frequencies to learn frequency-adaptive kernels of different sizes for better domain transformation.

\subsection{Adversarial attack and defense}
\label{sec:rconv:rw_aa}
Generally speaking, adversarial attack refers to deliberately designing inputs ($a.k.a.$ adversarial examples) to fool a trained network model and force it to produce wrong outputs. The purpose of adversarial attacks may vary under different scenarios \cite{carlini2019on}. In this paper, we focus on image recognition, where the goal of adversarial attacks is to cause misclassification.

Adversarial attack algorithms can be roughly classified into two categories based on whether gradient is adopted. Optimization-based attacks are by far the most popular methods. Given an input image and its associated ground truth label, these methods generate adversarial samples by computing the gradients according to the CNN architecture and the pre-defined loss function, such as cross-entropy loss, C\&W\cite{carlini2017cw}, \emph{etc.} Meanwhile, $\ell_1$~\cite{chen2018ead}, $\ell_2$~\cite{szegedy2014intriguing,carlini2017cw} and $\ell_\infty$~\cite{madry2018pgd,carlini2017provably} distortion metrics are commonly used to measure the budget of adversarial examples. In comparison, gradient-free attack methods are developed for the cases where the network architecture is unavailable. Some representative works can be found in \cite{chen2017zoo,uesato2018adversarial,ilyas2018blackbox}.

To defend against adversarial attacks, adversarial training \cite{athalye2018obfuscated} is one of the most popular and natural choices. It augments the training data by generating adversarial examples with certain attack methods. Though a re-trained model can deal with unseen data, it may still fail when facing adversarial samples generated by other attacking methods. This is because one model can hardly cover the entire input space by training with a certain number of searching steps. To narrow the gap, Deng \emph{et al.} \cite{dong2020adversarial} modelled the potential adversarial examples from the perspective of distribution. Other defence algorithms focus on designing better objective functions. Chen \emph{et al.} \cite{chen2019gce} proposed a novel loss function to neutralize the probability of wrong predictions and maximize the probability of right predictions. 

Different from the existing anti-attack methods, we focus on robust feature extraction of CNNs by proposing a new module. Our method is complementary to the existing adversarial attack defenders, and it can be readily used to improve the anti-attack performance of existing methods.

\subsection{Normalization layers}
\label{sec:rconv:norm}
Normalization layers are critical components of CNNs, aiming at reducing the internal covariate shift \cite{ioffe2015batch} between input distribution and output distribution. Batch Normalization (BN) \cite{ioffe2015batch} is the first work to mitigate this issue. It normalizes the whole batch by computing sample statistics (mean and standard deviation) during mini-batch based training. Layer Normalization (LN) \cite{jimmy2016layernorm} is designed to normalize all the activations of a single layer of a batch along the channel dimension, where it collects statistics from every unit within the layer. Besides, Instance Normalization (IN) \cite{ulyanov2016in} performs a BN-like computation to each sample, where a sample refers to a unit of the space spanned by the batch and channel axes. In addition, Group Normalization (GN) \cite{wu2018gn} improves BN by partitioning channels into groups and computing mean values and standard deviations for each group. 

In this paper, we propose a normalized soft-thresholding (NST) operator to deal with the covariate shift caused by different corruptions.

\section{Proposed Method}
\label{sec:rconv:method}
\subsection{Problem formulation}
Denote by $x\in\mathcal{R}^{H\times{W}\times{C}}$ an input image in the spatial domain and $y$ its ground truth label. A CNN model of $L$ layers can be regarded as a sequence of functions as follows:
\begin{equation}
\hat{y}={f}\circ{x}=f^{(L)}\circ{f}^{(L-1)}\circ\ldots\circ{f}^{(2)}\circ{f}^{(1)}\circ{x},
\end{equation}
where $f^{(i)}$ is the function of the $i$-th layer of the CNN model and its associated parameters are denoted as $\theta^{(i)}$, $i=1,2,\ldots,L$. We further denote the input and the output of $f^{(i)}$ by $x^{(i)}\in\mathcal{R}^{H^{(i)}_{in}\times{W}^{(i)}_{in}\times{C}^{(i)}_{in}}$ and $z^{(i)}\in\mathcal{R}^{H^{(i)}_{out}\times{W}^{(i)}_{out}\times{C}^{(i)}_{out}}$, respectively. Note that the output of the last layer, $z^{(L)}$, is the prediction of the ground truth label $y$, denoted by $\hat{y}$. Some loss function $\mathcal{J}_f(y,\hat{y})$, \emph{e.g.}, the cross-entropy loss, could be defined to measure the distance between $y$ and $\hat{y}$ and to update the CNN parameters.

In practice, the input image $x$ can be corrupted, and the corrupted sample can be written as
\begin{equation}
x'=x+\eta,
\end{equation}
where $\eta$ refers to the corruption, \emph{e.g.}, noise or the perturbation deliberately generated by a specific adversarial algorithm, aiming to enforce a wrong classification $y'=f(x')\neq{y}$ on corrupted sample. 

In this paper, we do not assume any specific distribution on $\eta$. It is expected that a robust CNN model could consistently make correct predictions for either the clean input image $x$ or its corrupted counterparts $x'$.
\subsection{Brief review of LST}
This work is inspired by LST-Net \cite{li2020lstnet}, and we briefly review it here. An LST contains three primitive transforms, including channel transform $T_c$, spatial transform $T_s$, and resize transform $T_r$. Both $T_c$ and $T_s$ work closely to reduce redundancies by converting the input into a frequency domain along the channel and spatial dimensions in order. Beginning with the DCT for training, they can be implemented by a learnable pointwise convolutional layer (PWConv) \cite{lin2014network} and a depthwise separable convolutional layer (DWConv) \cite{howard2017mobilenets}, respectively. Besides, their outputs are organized along the channel dimension for removal of noise and trivial features by soft-thresholding (ST). $T_s$ is always arranged right after $T_c$ to save parameters and computational cost \cite{li2020tcts}. Thus, the output features of $T_c$ (\emph{i.e.}, the input features of $T_s$) are expected to correspond to the weight of $T_c$ along the channel dimension, while it is assumed that high frequency signals always come after low frequency ones. $T_r$ is placed before or after the composition transform of $T_c$ and $T_s$ to obtain the desired output size by using a PWConv. 

We argue that the LST is restricted by its fixed kernel size from effective feature extraction for different frequency components. Besides, ST is less powerful to deal with corruptions at different levels. In this paper, we study both the problems and mitigate them with the proposed RConv-MK.
\subsection{The module structure}
\begin{wrapfigure}{r}{0.48\textwidth}
\vspace{-7mm}
\centering
	\begin{subfigure}{0.22\textwidth}
    \includegraphics[width=\linewidth]{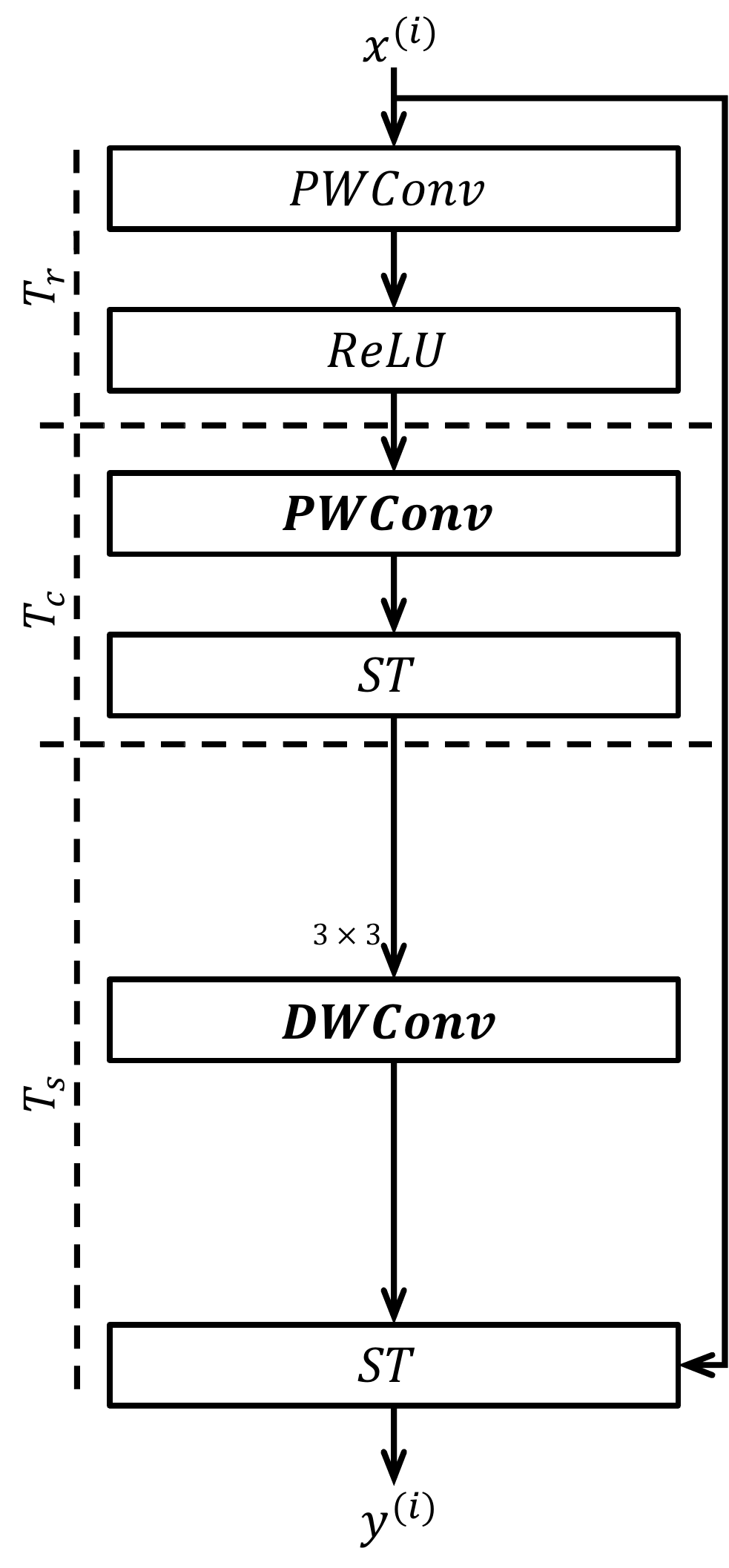}
    \caption{}
    \label{fig:rconv:lst_Ts}
    \end{subfigure}%
	\,
	\begin{subfigure}{0.22\textwidth}
    \includegraphics[width=\linewidth]{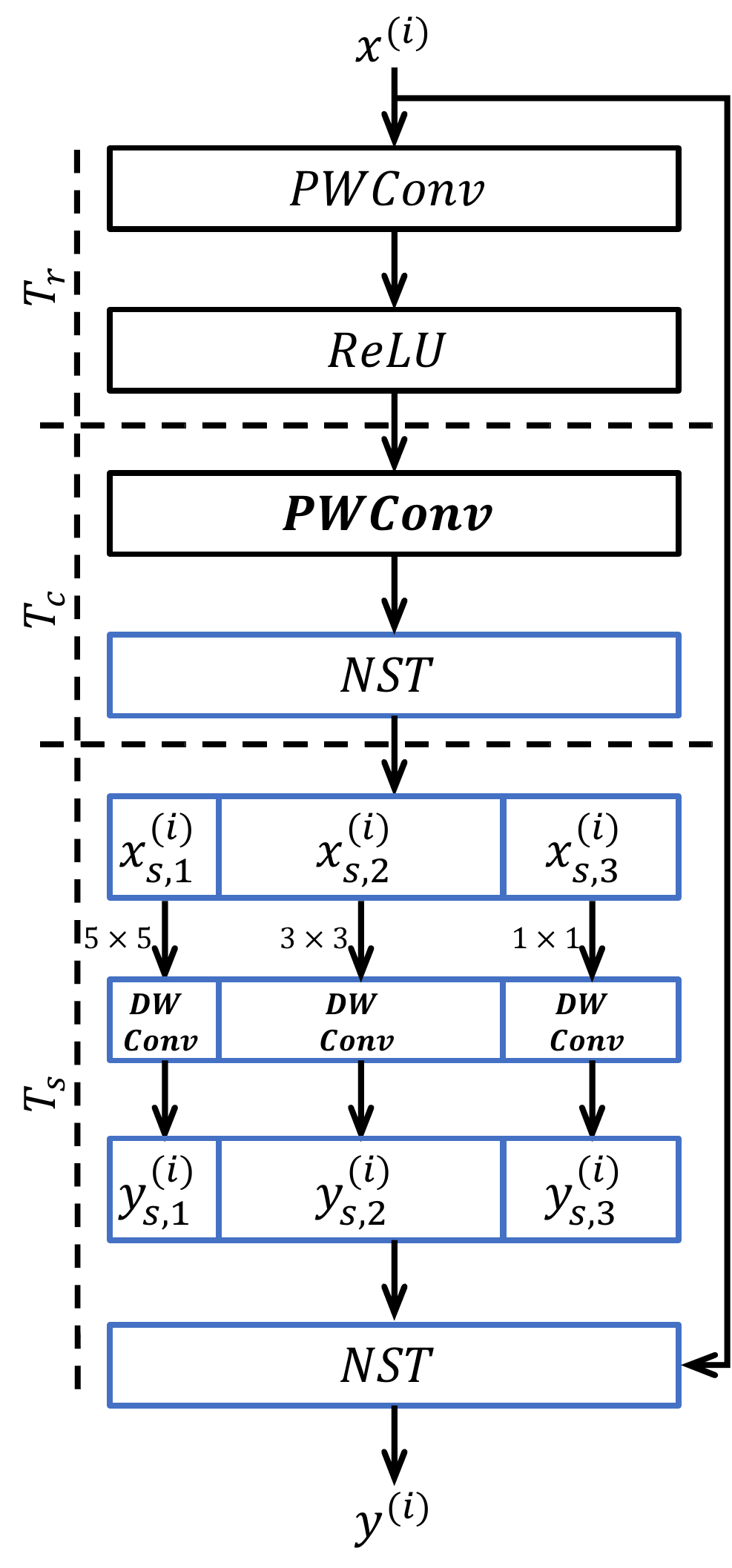}
    \caption{}
    \end{subfigure}%
    \vspace{-8pt}
    \caption{Structure comparison of (a) LST and (b) our RConv-MK. PWConv/DWConv in regular and bold font suggests random and DCT initialization of the associated weights, respectively.}
    \vspace{-10mm}
	\label{fig:rconv:structure}
\end{wrapfigure}

Figure \ref{fig:rconv:structure} presents the structure of LST \cite{li2020lstnet} and the proposed RConv-MK. Both of them consist of three primitive transforms: $T_c$, $T_s$ and $T_r$, where $T_s$ follows $T_c$, and $T_r$ is placed either before (as illustrated in Figure \ref{fig:rconv:structure}) or after (see our supplementary material) the other two transforms. The differences between LST and our RConv-MK are in two aspects. First, we design a spatial transform with multiple kernels to exploit the specific characteristics of different frequencies along the channel dimension. Second, we replace ST with NST to deal with corruptions at different levels.

\subsection{Spatial transform with multiple kernels}

Given the kernel size ${k}\times{k}$ and the expansion rate $a$ for DWConv, $T_s$ of LST \cite{li2020lstnet} repeatedly applies a convolutional kernel $\theta_{s}^{(i)}\in\mathcal{R}^{a^2\times1\times{k}\times{k}}$ to each channel of $x_s^{(i)}$, where the subscript $s$ suggests association to the  spatial transform. The learned weights of channel transform $T_c$ in LST maintain two important properties (which are also possessed by our RConv-MK). First, the transformed features of $T_c$ (\emph{i.e.}, the input features of $T_s$) are structured, where the low frequency features are placed at one end while the high frequency features are located at the other end in the channel dimension. Second, low frequency features are dense while high frequency features are sparse. Without loss of generality, in the remaining of this paper, we assume that features of $x_s^{(i)}$ (or $y_c^{(i)}$) are arranged from low to high frequencies along the channel dimension.

The $T_s$ in LST, unfortunately, loses the properties of $T_c$. The fixed kernel size (usually 3$\times$3 in most modern CNNs) may be too small to identify the genuine low frequency signals. Thus, some high frequency signals will be misclassified as low frequency ones due to the limited window size. Meanwhile, the fixed kernel size may not be suitable to efficiently process sparse high frequency signals (see our supplementary material for more discussions). 

\begin{figure}[t]
\begin{center}
    \begin{subfigure}{0.65\textwidth}
    \includegraphics[width=\linewidth]{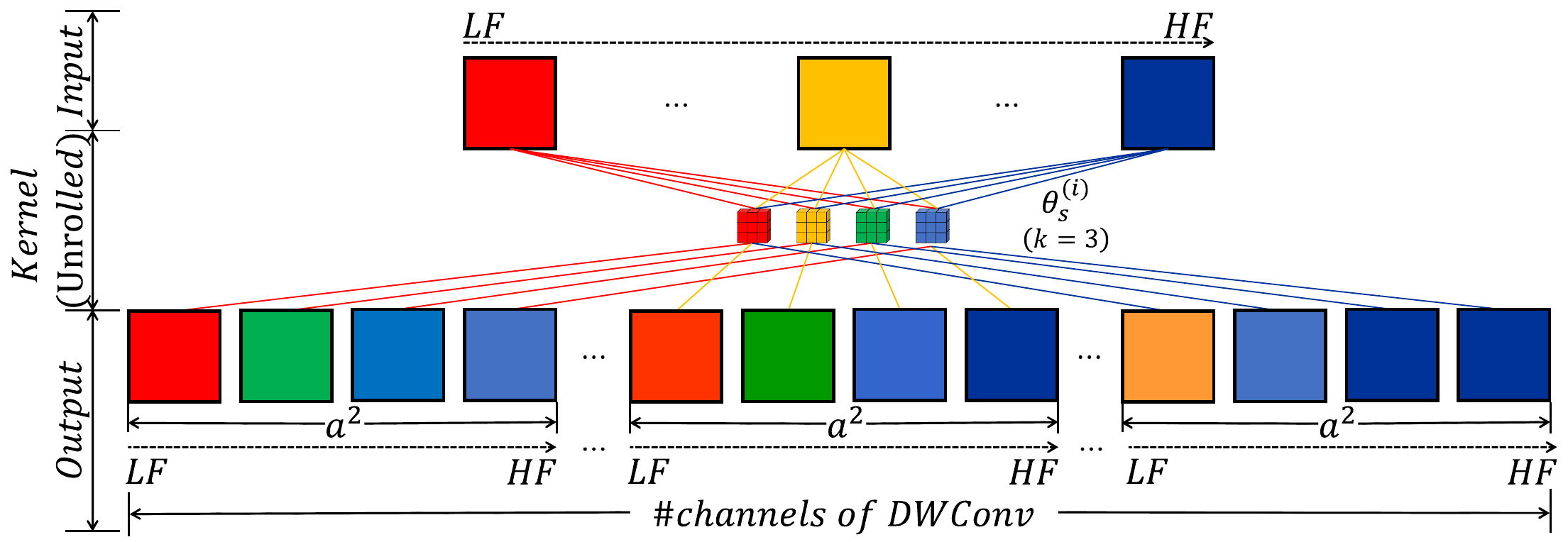}
    \caption{}
	\label{fig:rconv:lst_Ts}
    \end{subfigure}
    \\
	\begin{subfigure}{0.65\textwidth}
    \includegraphics[width=\linewidth]{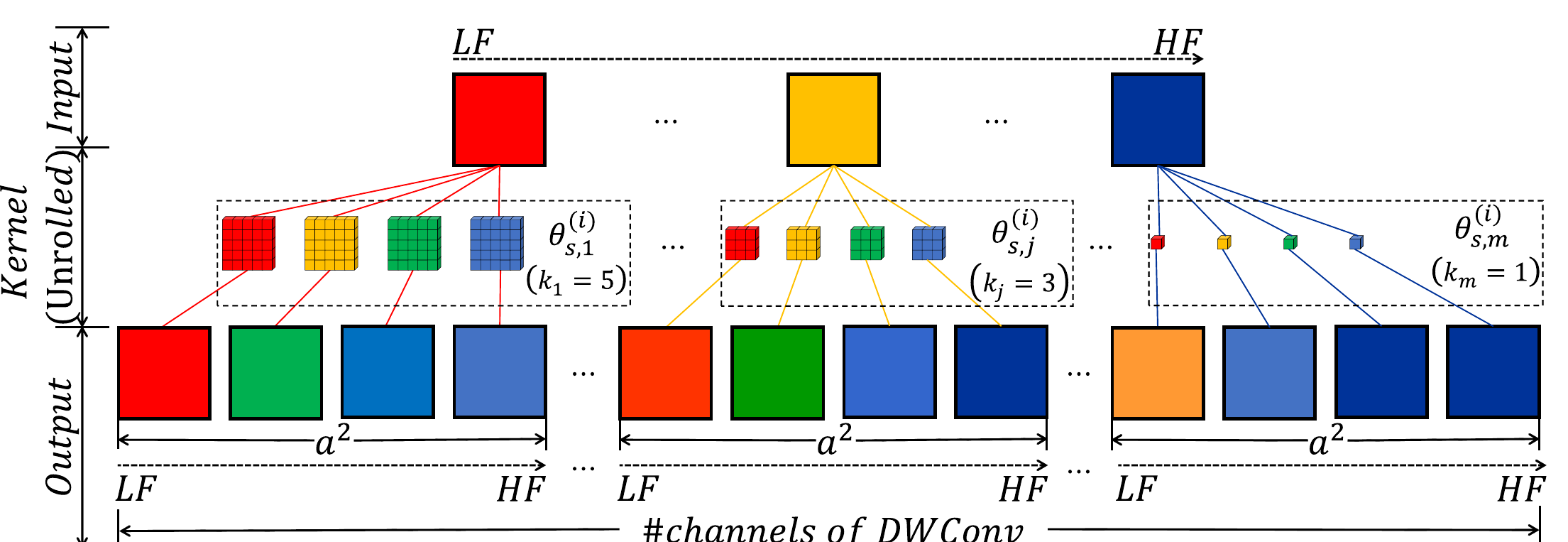}
    \caption{}
    \end{subfigure}
    \vspace{-8pt}
    \caption{Comparison of the spatial transform $T_s$ in (a) LST and (b) our RConv-MK, where ``LF'' and ``HF'' suggest the expected location of low and high frequency signals along the channel axis, respectively. In each case, the input channels are arranged by frequency and expanded for $a^2$ times. The $T_s$ of LST only adopts a single DWConv kernel, while our RConv-MK incorporates $m$ DWConv kernels of different sizes. The low-frequency components are computed with large kernels and high-frequency components with small kernels.}
	\label{fig:rconv:rconv_mk_Ts}
    \vspace{-8pt}
    \end{center}
\end{figure}

Figure~\ref{fig:rconv:rconv_mk_Ts} compares the design of $T_s$ in LST and our RConv-MK. 
Channels of the input and the output of $T_s$ are highlighted in prism colorset, where red means low frequency signals while blue means high frequency signals.
By using kernels of suitable sizes for signals of different frequencies, we can improve $T_s$ to produce better features for low frequency signals and accelerate the computation for high frequency ones. We leverage a set of $m$-kernels of different sizes and sort them by their kernel size in a descending order. The associated weights can be written as $\theta^{(i)}_s=\{\theta^{(i)}_{s,j}|\theta^{(i)}_{s,j}\in\mathcal{R}^{a^2\times1\times{k_j}\times{k_j}},j=1,\ldots,m\}$, satisfying $\forall{p}>{q}, k_p>k_q\geq1$, and $\exists{j}, k_j=k$. Accordingly, we partition $x^{(i)}_s$ into $m$ groups along the channel dimension to have $x^{(i)}_{s,1},\ldots,x^{(i)}_{s,m}$, where $x^{(i)}_{s,j}\in\mathcal{R}^{H^{(i)}_{in}\times{W}^{(i)}_{in}\times{C}^{(i)}_{s,j}}$, $\forall{j}=1,\ldots,m$. Obviously, $\sum_{j=1}^{m}{C}^{(i)}_{s,j}={C}^{(i)}_{s}$. $T_s$ of RConv-MK applies each $\theta^{(i)}_{s,j}$ to its related group of input channels $x^{(i)}_{s,j}$, $\forall{j}=1,\ldots,m$, so that low frequency signals are assigned with large kernels while high frequency ones are assigned with small kernels. In this way, we can make good use of signals of different frequencies in the feature domain. Finally, $T_s$ ends up with concatenating features of all $m$-groups along the channel dimension, producing the transformed features of $T_s$ in our RConv-MK.

\subsection{Normalized soft thresholding}
In LST, soft-thresholding is used to remove noise and trivial features. However, the threshold $\tau$ is determined manually based on the noise level in the corresponding feature domain. Mismatch of $\tau$ and features may cause performance drop. A large $\tau$ value may cut down useful cues, while noises still survive if a small $\tau$ is set. Actually, a fixed threshold threatens the robustness of a convolutional layer since the noise level of a corrupted sample may vary dramatically from one sample to another. It is highly desired to develop an adaptive thresholding scheme to remove noise and trivial features for robust convolution. We develop a normalized soft thresholding (NST) method to this end. Mathematically, NST first normalizes each sample $X_i$ ($i=1,\ldots,N$) in an $N-$sized mini-batch $X$ as
\begin{equation}
	\label{eq:rconv:ln}
	X_{LN,i}=(X_i-\mu({X_i}))/\sigma({X_i})
\end{equation}
where $\mu(\cdot)$ and $\sigma(\cdot)$ compute the mean and the standard deviation, respectively. In this way, corrupted samples at different levels are scaled to nearly the same level in an adaptive manner so that the normalized corrupted samples are expected to approach the distribution of their corresponding clean samples. In this sense, we are able to further mitigate the internal covariate shift of the mini-batch $X_{LN}$ by applying BN \cite{ioffe2015batch} to them, obtaining $X_{BN}$. The above procedures of NST can be easily implemented with a sequence of non-parametric LN plus a standard BN, which can be found in many existing toolkits \cite{abadi2016tensorflow,chen2016mxnet,paszke2017automatic}.

Finally, the corruptions in the normalized feature domain can be suppressed by a standard ST operation
\begin{equation}
    \label{eq:rconv:st}
        Y_{NST}=
        \begin{cases}
               sgn(X_{BN})(\left|X_{BN}\right|-\tau), & \left|X_{BN}\right|\geq\tau,\\
               0, & \ otherwise.
        \end{cases}
\end{equation}
where $\tau$ is the threshold and $Y_{NST}$ is the NST output of $X$. With the introduction of normalization in NST, we can easily set $\tau=10^{-4}$ in all experiments.

\begin{table}[t]
    \centering
    \caption{Methods for comparison in this paper.}
	\setlength{\tabcolsep}{1.8mm}
    \resizebox{0.9\textwidth}{!}{
    \begin{tabular}{cccccc}
        \toprule
        \multirow{2}{*}{Method} & Auxiliary & Noise & Receptive & Low frequency & High frequency  \\
        & branch & removal & field & kernel & kernel \\
        \midrule
        Conv2d & \xmark & N.A. & Uniform & N.A. & N.A. \\
        Conv2d-MK & \xmark & N.A. & Varied & N.A. & N.A. \\
        Conv2d+SE \cite{hu2018squeeze} & \cmark & N.A. & Uniform & N.A. & N.A.\\
        Conv2d+CBAM \cite{woo2018cbam} & \cmark & N.A. & Uniform & N.A. & N.A. \\
        LST \cite{li2020lstnet} & \xmark & ST & Uniform & N.A. & N.A. \\
        RConv-UK & \xmark & NST & Uniform & N.A. & N.A. \\
        RConv-RMK & \xmark & NST & Varied & Small & Large \\
        RConv-DMK & \xmark & N.A. & Varied & Large & Small \\
        RConv-LMK & \xmark & LN & Varied & Large & Small \\
        RConv-SMK & \xmark & ST & Varied & Large & Small \\
        RConv-MK & \xmark & NST & Varied & Large & Small \\
        \bottomrule
    \end{tabular}
	}
	\vspace{-8pt}
    \label{tab:rconv:compare_methods}
\end{table}

\subsection{Implementation details and complexity analysis}
By using the proposed spatial transform with multiple kernels and the NST operator, we are able to construct a robust convolutional layer, namely RConv-MK, for more reliable visual recognition. We expect that RConv-MK has almost the same overhead as LST at the cost of a negligible number of extra parameters for the same setting of $k$ and $a$. We set $m=3$ and fix $k_1=k+2$, $k_2=k$ and $k_3=1$ in this paper. Besides, we set ${C}^{(i)}_{s,2}$ to half of ${C}^{(i)}_{s}$ so that the kernel of size ${k}\times{k}$ will be computed with the majority of the input features. As of DWConv, when we have the same input shape and stride, the overhead is in proportion to the kernel size. Therefore, we have ${C}^{(i)}_{s,j}\propto{k}_j^{-2}$, $j=1,\ldots,m$. With the above settings, we encourage increase of input channels in high frequency (those convolved with kernels of smaller size) and decrease of input channels in low frequency (those convolved with kernels of larger size), which well matches recent findings that high frequency features are critical to the model generalization ability \cite{wang2020highfrequency} while low frequency features are rather vulnerable to adversarial attacks \cite{guo2020low}. We conduct a grid search to specify the proportion for some popular kernel sizes. In this paper, we set ${C}^{(i)}_{s,1}:{C}^{(i)}_{s,2}:{C}^{(i)}_{s,3}=1:3:2$ when $k=3$, and ${C}^{(i)}_{s,1}:{C}^{(i)}_{s,2}:{C}^{(i)}_{s,3}=1:2:1$ when $k=5$.

Compared to LST, the extra parameters of RConv-MK can be determined by $\{\theta^{(i)}_{s,j}|\theta^{(i)}_{s,j}\in\mathcal{R}^{a^2\times1\times{k_j}\times{k_j}},j=1,\ldots,m,k_j\neq{k}\}$. In modern CNN architectures, as $C_{in}^{(i)},C_{out}^{(i)}\gg{a},{k}$, the total number of parameters of RConv-MK are dominated by $T_r$ and $T_c$, which are the same as their counterparts of LST and proportional to $C_{in}^{(i)}\times{C}_{out}^{(i)}$. Weights of $T_s$ discussed in this paper only occupy a tiny fraction. Take the ResNet architecture \cite{he2016identity} as an example. Given $a=2$, there are only $a^2\times(k_1^2+k_3^2)=2^2\times(5^2+1^2)=104$ extra parameters in a RConv-MK. In contrast, $C_{in}^{(i)}$ and $C_{out}^{(i)}$ range in $[64,512]$. Approximately, the number of extra parameters only occupies 0.01\%$\sim$1\% of the total number of parameters in a RConv-MK.

\section{Experiments}
\label{sec:rconv:exp}
We perform extensive experiments to evaluate the robustness of the proposed RConv-MK to common types of corruptions and adversarial attacks. Ablation studies are also conducted to set the number of multiple kernels and the channel split in RConv-MK. All experiments are conducted on a 10-way NVIDIA RTX server. We use PyTorch~\cite{paszke2017automatic} for implementation. Due to page limit, results of the proposed RConv-MK under more CNN architectures and ablation study can be found in the \textbf{supplementary material}.

\subsection{Experiment setup and datasets}
\textbf{Methods for comparison}. As an alternative to Conv2d in a CNN, we compare the proposed RConv-MK with Conv2d and its variants as well as LST \cite{li2020lstnet}. Besides, we further test some variants of RConv-MK to better understand the roles of its different components. Table~\ref{tab:rconv:compare_methods} lists the competing methods and their attributes, including use of auxiliary branch, noise removal method, receptive field and low and high frequency kernel size.  Similar to RConv-MK, Conv2d-MK splits the input along channel dimension into $m$-groups, performs Conv2d with different kernel sizes for each group, and then concatenates the result of each group. We also combine two popular attention modules, \emph{i.e.}, SE~\cite{hu2018squeeze} and CBAM~\cite{woo2018cbam}, with Conv2d, denoted as Conv2d+SE and Conv2d+CBAM. As for the variants of RConv-MK, RConv-UK adopts a uniform kernel in $T_s$, RConv-RMK reverses the order of kernels, RConv-LMK replaces NST by LN, RConv-SMK substitutes NST with ST, and RConv-DMK removes all NST operators.

\textbf{Tasks and datasets}. We compare the competing methods on three tasks: visual recognition on corrupted images, white-box adversarial attacks, as well as recognition on clean images. The ImageNet-C \cite{hendrycks2019robustness} dataset is employed to evaluate the robustness of each method to common corruptions. CIFAR-10/100 \cite{krizhevsky2009learning} are used for the evaluation under white-box adversarial attacks. The ImageNet~\cite{deng2009imagenet} dataset is employed for evaluating classification performance on clean images. In addition, we also employ the MS-COCO~\cite{lin2014mscoco} dataset to evaluate the proposed method for object detection and instance segmentation. On each dataset, we closely follow the standard experimental settings for fair comparison. Details can be found in our supplementary material.

\subsection{Evaluation on images with corruptions}
To study the generalization ability of models trained with clean images to various corruptions, ImageNet-C~\cite{hendrycks2019robustness} is constructed by applying 19 types of distinct corruptions to the validation set of ImageNet \cite{deng2009imagenet}. The mean corruption error (mCE) is used as the criteria (the lower the better) for performance evaluation. We build up CNNs of different competing methods under ResNet-50. The best snapshot of each method on ImageNet \cite{deng2009imagenet} is used for comparison. 

\begin{table}[tb]
    \centering
    \caption{Comparison of robustness to common corruptions under ResNet-50 architecture on ImageNet-C.}
	\setlength{\tabcolsep}{2.0mm}
    \resizebox{0.6\textwidth}{!}{
	    \begin{tabular}{ccc}
             \toprule Method & Top-1/5 E. R. ($\%$) & mCE \\
             \midrule
             Conv2d & 23.85/7.13 & 77.01 \\
             Conv2d+SE \cite{hu2018squeeze} & 23.14/6.70 & 74.47 \\
             Conv2d+CBAM \cite{woo2018cbam} & 22.98/6.68 & 72.56 \\
             Conv2d-MK & 24.96/7.51 & 77.17 \\
             LST \cite{li2020lstnet} & 22.78/6.66 & 70.54 \\
             RConv-LMK & 22.76/7.05 & 70.34 \\ 
             RConv-SMK & 22.98/6.64 & 70.80 \\ 
             RConv-DMK & 23.31/6.88 & 70.93 \\ 
             RConv-RMK & 23.10/6.80 & 70.81 \\
             RConv-UK & 22.59/6.58 & 69.79 \\
             RConv-MK & \textbf{22.22}/\textbf{6.32} & \textbf{67.91} \\
             \bottomrule
        \end{tabular}
	}
	\vspace{-8pt}
	\label{tab:rconv:imagenet_c_resnet50}
\end{table}

Table \ref{tab:rconv:imagenet_c_resnet50} shows the best top-1/5 error rates on clean ImageNet and the corresponding mCE values on ImageNet-C. 
RConv-MK obtains lower mCE than all its competitors. It significantly reduces the mCE of the baseline Conv2d by 9.10$\%$. 
Besides, in terms of corruption suppression methods, RConv-MK $>$ RConv-LMK or RConv-SMK $>$ RConv-DMK. This shows both LN and ST improve the model robustness to common corruptions, while the proposed NST can produce more robust results. Besides, RConv-MK $>$ LST $>$ RConv-SMK. This suggests our NST is more robust to unseen corruptions than ST for multiple kernels as NST normalizes the features into the same range for noise and redundancy removal. In contrast, ST considers the amplitude values only. Though the adoption of multiple kernels helps feature extraction by frequency, RConv-SMK may also increase the amplitude change in some supporting frequencies of a corrupted image, which weakens its robustness. 
Furthermore, when it comes to the arrangement of multiple kernels, RConv-MK (normal order) $>$ RConv-UK (uniform kernel) $>$ RConv-RMK (reversed order). This suggests signals of different frequencies are sensible to the kernel size. Mismatches make it even worse than using a uniform kernel. 
In addition, it is critical to group and concatenate channels for multiple kernels in the frequency domain. We see RConv-MK $>$ RConv-UK in frequency domain while Conv2d-MK $<$ Conv2d because signals in the spatial domain are not well structured.

\subsection{Evaluation on adversarial attacks}
Our RConv-MK is developed from the perspective of network architecture so that comparison against existing adversarial training algorithms lies out of our main focus. Actually, our method is complementary to these methods in practice. Below, we compare the robustness of RConv-MK and its competing network building blocks to adversarial attacks on CIFAR-10/100. We build models under WRN34-10 (results under ResNet-18 are presented in our supplementary material). We conduct adversarial training of each model on each dataset under the $\ell_\infty$ PGD attack for 100 epochs with common hyper-parameter settings. Specifically, the perturbation size is $\epsilon=8/255$, step size is $\eta=2/255$, and number of steps is 10. Learning rate starts at 0.1 and is reduced by a factor of 10 after 75, 90 and 100 epochs, respectively. We fix the batch size as 128 and weight decay as 0.0002. We test each trained model under untargeted white-box attacks with five representative anti-attack algorithms, including FGSM~\cite{goodfellow2015fgsm}, PGD~\cite{madry2018pgd}, FFGSM~\cite{wong2020ffgsm}, ODI~\cite{tashiro2020odi} and AWP~\cite{wu2020awp}. We use the official implementation of both ODI and AWP, and we exploit advertorch \cite{ding2019advertorch} of the rest.

\begin{table}[t]
    \centering
    \caption{Results (robust accuracy, $\%$) by different methods under untargeted white-box attacks on CIFAR-10/100.}
    \setlength{\tabcolsep}{1.5mm}
    \resizebox{\textwidth}{!}{
        \begin{tabular}{ccccccccccccc}
             \toprule
             \multirow{2}{*}{Attacks} & \multirow{2}{*}{Dataset} & \multirow{2}{*}{Conv2d} & Conv2d+ & Conv2d+ & Conv2d-  & LST  & RConv- & RConv- & RConv- & RConv- & RConv- & RConv- \\
             & & & SE \cite{hu2018squeeze} & CBAM \cite{woo2018cbam} & MK & \cite{li2020lstnet} &  LMK & SMK & DMK & RMK & UK & MK \\
             \midrule
			 \multirow{2}{*}{FFGSM \cite{wong2020ffgsm}} & C10 & 60.78 & 60.75 & 60.49 & 60.87 & 62.80 & 64.06 & 64.18 & 62.32 & 63.60 & 64.05 & \textbf{64.55} \\
			 & C100  & 32.15 & 32.04 & 32.35 & 31.12 & 34.08 & 32.48 & 33.91 & 33.84 & 33.66 & 34.19 & \textbf{34.55} \\
			 \multirow{2}{*}{FGSM \cite{goodfellow2015fgsm}} & C10 & 57.31 & 56.70 & 56.68 & 57.63 & 59.02 & 60.16 & 60.18 & 57.71 & 58.56 & 59.86 & \textbf{60.67} \\
			 & C100  & 29.22 & 28.98 & 29.62 & 28.29 & 30.25 & 30.12 & 30.82 & 29.90 & 30.04 & 31.07 & \textbf{31.50} \\
			 \multirow{2}{*}{PGD \cite{madry2018pgd}} & C10 & 47.05 & 46.51 & 46.89 & 47.04 & 50.46 & 50.92 & 50.95 & 49.60 & 50.24 & 51.38 & \textbf{52.64} \\
			 & C100  & 24.00 & 23.34 & 24.05 & 22.94 & 24.42 & 24.90 & 25.42 & 25.06 & 25.39 & 25.51 & \textbf{26.63} \\
			 \multirow{2}{*}{ODI \cite{tashiro2020odi}} & C10 & 45.94 & 45.36 & 45.62 & 45.64 & 48.21 & 49.19 & 50.53 & 48.08 & 49.23 & 50.29 & \textbf{51.05} \\
			 & C100  & 22.85 & 22.20 & 22.84 & 21.56 & 24.37 & 23.81 & 24.94 & 24.10 & 24.93 & 25.03 & \textbf{25.39}  \\
			 \multirow{2}{*}{AWP \cite{wu2020awp}} & C10 & 56.17 & 56.11 & 55.90 & 56.21 & 56.22 & 57.76 & 57.81 & 56.89 & 57.32 & 57.59  & \textbf{58.22} \\
			 & C100  & 28.80 & 28.62 & 28.86 & 27.61 & 29.03 & 29.03 & 29.10 & 29.09 & 29.07 & 29.14 & \textbf{29.46} \\
             \bottomrule
        \end{tabular}
    }
    \vspace{-8pt}
    \label{tab:rconv:whitebox_attack}
\end{table}

Table~\ref{tab:rconv:whitebox_attack} presents the accuracy obtained by different methods. One can have the following findings. First, the proposed RConv-MK outperforms all its competitors for adversarial attacks. Second, the attention modules, including Conv2d+SE and Conv2d+CBAM, have almost the same performance as the baseline under various untargeted white-box attacks. According to their definition, both attention modules pay more attention to local patterns while they suppress trivial features for image recognition. Although such kind of mechanism is helpful to the recognition of clean images, it may hurt the backbone model under adversarial attacks because the corrupted local patterns have a bigger chance to impose uncorrected excitation on the target features. Third, we see that RConv-MK $>$ RConv-LMK $>$ RConv-SMK $>$ RConv-DMK on CIFAR-10 and RConv-MK $>$ RConv-SMK $>$ RConv-DMK $>$ RConv-LMK on CIFAR-100. ST improves the robustness to adversarial attacks as it actually plays a role of gradient mask under adversarial attacks. LN shows competitive performance on CIFAR-10 but it performs poorly on CIFAR-100. This may result from the over-fitting problem of LN in adversarial training. With the increase of categories, the decision boundaries are expected to be less smooth in the feature space shaped by LN. Therefore, the model becomes vulnerable to unseen adversarial samples during test. Fourth, the arrangement of multiple kernels also matters in adversarial attacks. We can see that RConv-MK (normal order) $>$ RConv-UK (uniform kernel) $>$ RConv-RMK (reversed order). Fifth, Conv2d-MK always obtains worse results than the baseline due to its poor structure in spatial domain for channel split and concatenation. In contrast, our RConv-MK improves RConv-UK under all attacks as the channel operations are conducted in a well-structured space.

\begin{table}[t]
    \centering
    \caption{Results (error rates, $\%$) of RConv-MK under ResNet architecture on ImageNet.}
	\setlength{\tabcolsep}{2.0mm}
    \resizebox{0.6\textwidth}{!}{
        \begin{tabular}{cccc}
         \toprule
         Depth & Method & Param/FLOPs & Top-1/Top-5 \\
         \midrule
         \multirow{3}{*}{18} & Conv2d & 11.69M/1.81G & 30.24/10.92 \\
          & LST \cite{li2020lstnet} & \textbf{8.03M}/\textbf{1.48G} & 26.55/8.59 \\
          & RConv-MK & \textbf{8.03M}/\textbf{1.48G} & \textbf{26.26}/\textbf{8.48} \\
          \midrule
          \multirow{3}{*}{34} & Conv2d  & 21.79M/3.66G & 26.70/8.58 \\
          & LST \cite{li2020lstnet} & \textbf{13.82M}/\textbf{2.56G} & 23.92/7.24 \\
          & RConv-MK & \textbf{13.82M}/\textbf{2.56G} & \textbf{23.54}/\textbf{6.99} \\
          \midrule
          \multirow{3}{*}{50} & Conv2d & 25.56M/4.09G & 23.85/7.13 \\
          & LST \cite{li2020lstnet} & \textbf{23.33M}/\textbf{4.05G} & 22.78/6.66 \\
          & RConv-MK & \textbf{23.33M}/\textbf{4.05G} & \textbf{22.22}/\textbf{6.32} \\
          \midrule
          \multirow{3}{*}{101} & Conv2d & 44.55M/7.80G & 22.63/6.44 \\
          & LST \cite{li2020lstnet} & \textbf{42.36M}/\textbf{7.75G} & 21.63/5.94 \\
          & RConv-MK & \textbf{42.36M}/\textbf{7.75G} & \textbf{21.41}/\textbf{5.93} \\
		  \midrule
          \multirow{3}{*}{152} & Conv2d & 60.19M/11.51G & 21.69/5.94 \\
          & LST \cite{li2020lstnet} & \textbf{58.02M}/\textbf{11.46G} & 20.02/5.26 \\
          & RConv-MK & \textbf{58.02M}/\textbf{11.46G} & \textbf{19.77}/\textbf{5.15} \\
         \bottomrule
        \end{tabular}
	}
	\vspace{-8pt}
    \label{tab:rconv:imagenet_resnet}
\end{table}

\subsection{Evaluation on clean images}
We further study the performance of RConv-MK on clean images. We evaluate it on tasks of image recognition, object detection and instance segmentation.

\textbf{Image recognition}. The ImageNet \cite{deng2009imagenet} dataset is used to evaluate the performance of RConv-MK on image recognition with clean images.  We construct the models under the ResNet \cite{he2016identity} architecture and train/test them with the standard settings. Table \ref{tab:rconv:imagenet_resnet} shows the results. One can see that RConv-MK reduces the top-1/5 error rates of Conv2d by 1.22\%$\sim$3.98\% with less cost, and those of LST by 0.2\%$\sim$0.5\% at almost the same cost. This validates that RConv-MK can also improve the generalization performance of a CNN on clean images. 

We also compare RConv-MK with the DCTNet~\cite{xu2020dctnet} with 64 input channels under the same ResNet-50 architecture on ImageNet.  RConv-MK can reduce the top-1/5 error rates of DCTNet from 22.84$\%$/6.53$\%$ to 22.22$\%$/6.32$\%$. Besides, RConv-MK runs at 27.78 FPS (including data loading and pre-processing with single CPU thread plus computation on GPU), faster than DCTNet by 3.39 FPS. Though DCTNet can reduce the latency of data transmission to some extent by performing DCT sequentially on CPU, we note that the cost is still expensive (even with the support of advanced CPU instructions, \emph{e.g.}, AVX512). 

\begin{table}[t]
    \centering
    \caption{Object detection results ($\%$) of RConv-MK on MS-COCO validation set.}
	\setlength{\tabcolsep}{2.0mm}
    \resizebox{0.6\textwidth}{!}{
		    \begin{tabular}{ccccc}
				 \toprule
				 Detector & Backbone Method & ${mAP}$ & ${AP}_{50}$ & ${AP}_{75}$  \\
				 \midrule
				 \multirow{3}{*}{\begin{minipage}{0.8in}\centering Faster R-CNN \cite{ren2015faster}\end{minipage}} & Conv2d & 37.4 & 58.1 & 40.4 \\
                                                                         & LST \cite{li2020lstnet} & 40.8 & 62.2 & 44.3 \\
													                     & RConv-MK  & \textbf{41.3} & \textbf{62.6} & \textbf{45.0} \\ 
				 \midrule
				 \multirow{3}{*}{\begin{minipage}{0.8in}\centering RetinaNet \cite{lin2017retinanet}\end{minipage}} & Conv2d & 36.5 & 55.4 & 39.1 \\
                                                                         & LST \cite{li2020lstnet} & 38.7 & 58.5 & 41.7 \\
													                     & RConv-MK  & \textbf{39.4} & \textbf{60.0} & \textbf{42.0} \\
				 \midrule
				 \multirow{3}{*}{\begin{minipage}{0.8in}\centering FCOS \cite{tian2019fcos}\end{minipage}} & Conv2d & 36.6 & 55.7 & 38.8 \\
                                                                         & LST \cite{li2020lstnet} & 38.8 & 58.7 & 41.5 \\
													                     & RConv-MK  & \textbf{39.6} & \textbf{60.0} & \textbf{42.2} \\
				 \midrule
				 \multirow{3}{*}{\begin{minipage}{0.8in}\centering Mask R-CNN \cite{he2017maskrcnn}\end{minipage}} & Conv2d & 38.2 & 	58.8 & 41.4 \\
                                                                         & LST \cite{li2020lstnet}   & 41.3 & 62.5 & 45.0 \\
													                     & RConv-MK  & \textbf{41.8} & \textbf{63.3} & \textbf{45.9} \\
				 \midrule
				 \multirow{3}{*}{\begin{minipage}{0.8in}\centering Cascade Mask R-CNN \cite{cai2019cascadercnn}\end{minipage}}  & Conv2d & 41.2 & 59.4 & 45.0 \\
                                                                         & LST \cite{li2020lstnet} & 43.9 & 62.6 & 47.9 \\
													                     & RConv-MK  & \textbf{44.4} & \textbf{63.0} & \textbf{48.4} \\
				 \bottomrule
			\end{tabular}
	}
	\vspace{-3mm}
    \label{tab:rconv:objectdetection}
\end{table}

\begin{table}[t]
    \centering
    \caption{Instance segmentation results ($\%$) of RConv-MK on MS-COCO validation set.}
	\setlength{\tabcolsep}{2.0mm}
    \resizebox{0.6\textwidth}{!}{
		    \begin{tabular}{cccccc}
				 \toprule
				 Detector & Backbone Method & ${mAP}$ & ${AP}_{50}$ & ${AP}_{75}$  \\
				 \midrule
				 \multirow{3}{*}{\begin{minipage}{0.8in}\centering Mask R-CNN \cite{he2017maskrcnn}\end{minipage}} & Conv2d & 34.7 & 55.7 & 37.2 \\
                                                                         & LST \cite{li2020lstnet}  & 37.1 & 59.3 & 39.4 \\
													                     & RConv-MK & \textbf{37.6} & \textbf{59.9} & \textbf{40.2} \\
				 \midrule
				 \multirow{3}{*}{\begin{minipage}{0.8in}\centering Cascade Mask R-CNN \cite{cai2019cascadercnn}\end{minipage}} & Conv2d & 35.9 & 56.6 & 38.4 \\
                                                                         & LST \cite{li2020lstnet}  & 38.1 & 59.7 & 40.9 \\
													                     & RConv-MK  & \textbf{38.6} & \textbf{60.2} & \textbf{41.5}  \\
				 \bottomrule
			\end{tabular}
	}
	\vspace{-8pt}
    \label{tab:rconv:instancesegmentation}
\end{table}

\textbf{Object detection and instance segmentation}. We test the performance of RConv-MK for object detection and instance segmentation on MS-COCO~\cite{lin2014mscoco} by using representative object detectors such as Faster R-CNN \cite{ren2015faster} and Mask R-CNN \cite{he2017maskrcnn}, \emph{etc}.  Table \ref{tab:rconv:objectdetection} and Table \ref{tab:rconv:instancesegmentation} demonstrate the object detection and instance segmentation results on MS-COCO validation set, respectively. RConv-MK achieves better mAP than Conv2d and LST on both tasks. Among all object detectors, RConv-MK improves the mAP of Conv2d by 2.9\%$\sim$3.9\%, while it boosts the mAP of LST by 0.5\%$\sim$1.7\%. On instance segmentation, RConv-MK outperforms Conv2d by 2.7\%$\sim$2.9\% and LST by 0.5\% with Mask R-CNN and Cascade Mask R-CNN.

\begin{figure}[t]
   \begin{center}
   	\includegraphics[width=.85\linewidth]{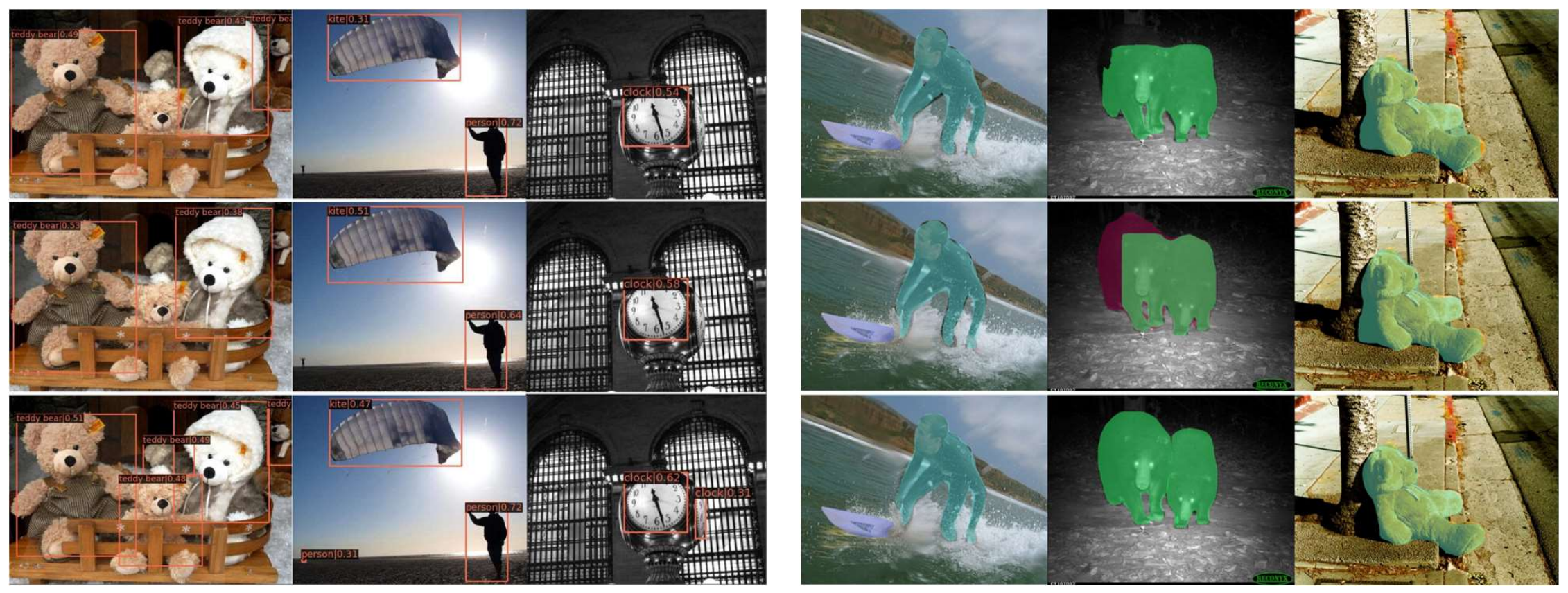}	
    \end{center}
        \vspace{-2pt}
       \caption{Comparisons of object detection results based on FCOS~\cite{tian2019fcos} and instance segmentation results based on Mask R-CNN~\cite{he2017maskrcnn} (from top to bottom: Conv2d, LST~\cite{li2020lstnet} and our RConv-MK).}
    	\label{fig:rconv:obj_det_vis}
    \vspace{-4mm}
    \end{figure}

Figure \ref{fig:rconv:obj_det_vis} (left) presents some visualization comparisons of object detection with FCOS. Compared with Conv2d and LST, though our RConv-MK leverages intermediate features the least in dimension for fusion, it obtains better detection results on challenging objects, for example, with severe occlusion (see the teddy bears in the left image), various sizes (see the persons in the middle image), as well as different levels of out-of-plane rotation (see the clocks in the right image). Such results demonstrate the robustness of RConv-MK in feature learning. 
In Figure \ref{fig:rconv:obj_det_vis} (right), we visualize instance segmentation results using Mask R-CNN. With the improved spatial transform, the proposed RConv-MK in the last row generates more precise segmentation results than Conv2d and LST in the middle two rows. For example, in the middle column, RConv-MK successfully segments the back of the left bear in the shadow, while Conv2d misses this part and LST misclassifies it as another object. 
\section{Conclusion}
\label{sec:rconv:conclusion}
In this paper, we proposed a robust alternative of Conv2d layer, namely RConv-MK, as a reliable feature extractor for visual recognition with corrupted images and adversarial samples. RConv-MK was designed with a set of kernels of different sizes so that they could be flexibly applied to the input features of different frequencies to exploit their specific characteristics. A normalized soft thresholding (NST) operator was then introduced to adaptively suppress the effect of different corruptions at different levels by using a uniform threshold. RConv-MK can be easily and efficiently implemented by the existing toolkits. Extensive experiments on corrupted images, adversarial samples as well as clean images validated the effectiveness of RConv-MK under popular CNN architectures.



\clearpage
{\small
\bibliographystyle{splncs04}
\bibliography{egbib}

\begin{thebibliography}{10}
\providecommand{\url}[1]{\texttt{#1}}
\providecommand{\urlprefix}{URL }
\providecommand{\doi}[1]{https://doi.org/#1}

\bibitem{abadi2016tensorflow}
Abadi, M., Barham, P., Chen, J., Chen, Z., Davis, A., Dean, J., Devin, M.,
  Ghemawat, S., Irving, G., Isard, M., Kudlur, M., Levenberg, J., Monga, R.,
  Moore, S., Murray, D.G., Steiner, B., Tucker, P., Vasudevan, V., Warden, P.,
  Wicke, M., Yu, Y., Zheng, X.: {TensorFlow}: A system for large-scale machine
  learning. In: Symp. Oper. Syst. Des. Implement. (2016)

\bibitem{athalye2018obfuscated}
Athalye, A., Carlini, N., Wagner, D.: Obfuscated gradients give a false sense
  of security: Circumventing defenses to adversarial examples. In: Int. Conf.
  Mach. Learn. (2018)

\bibitem{jimmy2016layernorm}
Ba, J.L., Kiros, J.R., Hinton, G.E.: Layer normalization. arXiv preprint
  arXiv:1607.06450  (2016)

\bibitem{cai2019cascadercnn}
Cai, Z., Vasconcelos, N.: Cascade {R-CNN}: High quality object detection and
  instance segmentation. IEEE Trans. Pattern Anal. Mach. Intell.  (2019).
  \doi{10.1109/TPAMI.2019.2956516}

\bibitem{carlini2019on}
Carlini, N., Athalye, A., Papernot, N., Brendel, W., Rauber, J., Tsipras, D.,
  Goodfellow, I., Madry, A., Kurakin, A.: On evaluating adversarial robustness.
  arXiv preprint arXiv:1902.06705  (2019)

\bibitem{carlini2017provably}
Carlini, N., Katz, G., Barrett, C., Dill, D.L.: Provably minimally-distorted
  adversarial examples. arXiv preprint arXiv:1709.10207  (2017)

\bibitem{carlini2017cw}
Carlini, N., Wagner, D.: Towards evaluating the robustness of neural networks.
  In: IEEE Symp. Secur. Priv. (2017)

\bibitem{chen2019gce}
Chen, H., Liang, J., Chang, S., Pan, J., Chen, Y., Wei, W., Juan, D.: Improving
  adversarial robustness via guided complement entropy. In: Int. Conf. Comput.
  Vis. (2019)

\bibitem{chen2018ead}
Chen, P., Sharma, Y., Zhang, H., Yi, J., Hsieh, C.: {EAD}: {Elastic-Net}
  attacks to deep neural networks via adversarial examples. In: AAAI (2018)

\bibitem{chen2017zoo}
Chen, P., Zhang, H., Sharma, Y., Yi, J., Hsieh, C.: {ZOO}: Zeroth order
  optimization based black-box attacks to deep neural networks without training
  substitute models. arXiv preprint arXiv:1708.03999  (2017)

\bibitem{chen2016mxnet}
Chen, T., Li, M., Li, Y., Lin, M., Wang, N., Wang, M., Xiao, T., Xu, B., Zhang,
  C., Zhang, Z.: {MXNet}: A flexible and efficient machine learning library for
  heterogeneous distributed systems. In: Adv. Neural Inform. Process. Syst.
  Worksh. (2016)

\bibitem{deng2009imagenet}
Deng, J., Dong, W., Socher, R., Li, L.J., Li, K., Fei-Fei, L.: {ImageNet}: A
  large-scale hierarchical image database. In: IEEE Conf. Comput. Vis. Pattern
  Recog. IEEE (2009)

\bibitem{ding2019advertorch}
Ding, G.W., Wang, L., Jin, X.: {AdverTorch} v0.1: An adversarial robustness
  toolbox based on pytorch. arXiv preprint arXiv:1902.07623  (2019)

\bibitem{dong2020adversarial}
Dong, Y., Deng, Z., Pang, T., Su, H., Zhu, J.: Adversarial distributional
  training for robust deep learning. In: Adv. Neural Inform. Process. Syst.
  (2020)

\bibitem{fan2020scn}
Fan, Y., Yu, J., Liu, D., Huang, T.S.: Scale-wise convolution for image
  restoration. In: AAAI (2020)

\bibitem{franzen2018image}
Franzen, F.: Image classification in the frequency domain with neural networks
  and absolute value {DCT}. In: Int. Conf. Image Signal Process. (2018)

\bibitem{furuta2019fcn}
Furuta, R., Inoue, N., Yamasaki, T.: Fully convolutional network with
  multi-step reinforcement learning for image processing. In: AAAI (2019)

\bibitem{goodfellow2015fgsm}
Goodfellow, I., Shlens, J., Szegedy, C.: Explaining and harnessing adversarial
  examples. In: Int. Conf. Learn. Represent. (2015)

\bibitem{guo2020low}
Guo, C., Frank, J.S., Weinberger, K.Q.: Low frequency adversarial perturbation.
  In: Uncertain. Artif. Intell. (2020)

\bibitem{he2017maskrcnn}
He, K., Gkioxari, G., Doll{\'a}r, P., Grishick, R.: Mask {R-CNN}. In: Int.
  Conf. Comput. Vis. (2017)

\bibitem{he2016identity}
He, K., Zhang, X., Ren, S., Sun, J.: Identity mappings in deep residual
  networks. In: Eur. Conf. Comput. Vis. Springer (2016)

\bibitem{hendrycks2019robustness}
Hendrycks, D., Dietterich, T.: Benchmarking neural network robustness to common
  corruptions and perturbations. Int. Conf. Learn. Represent.  (2019)

\bibitem{hossain2019distortion}
Hossain, T., Teng, S.W., Zhang, D., Lim, S., Lu, G.: Distortion robust image
  classification using deep convolutional neural network with discrete cosine
  transform. In: IEEE Int. Conf. Image Process. (2019)

\bibitem{howard2017mobilenets}
Howard, A.G., Zhu, M., Chen, B., Kalenichenko, D., Wang, W., Weyand, T.,
  Andreetto, M., Adam, H.: {MobileNets}: Efficient convolutional neural
  networks for mobile vision applications. arXiv preprint arXiv:1704.04861
  (2017)

\bibitem{hu2018squeeze}
Hu, J., Shen, L., Sun, G.: Squeeze-and-excitation networks. In: IEEE Conf.
  Comput. Vis. Pattern Recog. (2018)

\bibitem{ilyas2018blackbox}
Ilyas, A., Engstrom, L., Athalye, A., Lin, J.: Black-box adversarial attacks
  with limited queries and information. In: Int. Conf. Mach. Learn. (2018)

\bibitem{ioffe2015batch}
Ioffe, S., Szegedy, C.: Batch normalization: Accelerating deep network training
  by reducing internal covariate shift. In: Int. Conf. Mach. Learn. (2015)

\bibitem{krizhevsky2009learning}
Krizhevsky, A., Hinton, G.E.: Learning multiple layers of features from tiny
  images. Tech. Rep. TR-2009, University of Toronto (2009)

\bibitem{li2020lstnet}
Li, L., Wang, K., Li, S., Feng, X., Zhang, L.: {LST-Net}: Learning a
  convolutional neural network with a learnable sparse transform. In: Eur.
  Conf. Comput. Vis. (2020)

\bibitem{li2020tcts}
Li, L., Wang, K., Li, S., Feng, X., Zhang, L.: Remarks on {Tc} and {Ts} (2020),
  \url{https://github.com/lld533/LST-Net/blob/master/Remarks_on_Tc_and_Ts.txt}

\bibitem{lin2014network}
Lin, M., Chen, Q., Yan, S.: Network in network. In: Int. Conf. Learn.
  Represent. (2014)

\bibitem{lin2017retinanet}
Lin, T.Y., Goyal, P., Girshick, R., He, K., Doll{\'a}r, P.: Focal loss for
  dense object detection. In: Int. Conf. Comput. Vis. (2017)

\bibitem{lin2014mscoco}
Lin, T.Y., Maire, M., Belongie, S., Bourdev, L., Girshick, R., Hays, J.,
  Perona, P., Ramanan, D., Zitnick, C.L., Doll{\'a}r, P.: {Microsoft COCO}:
  Common objects in context. In: Eur. Conf. Comput. Vis. (2014)

\bibitem{madry2018pgd}
Madry, A., Makelov, A., Schmidt, L., Tsipras, D., Vladu, A.: Towards deep
  learning models resistant to adversarial attacks. In: Int. Conf. Learn.
  Represent. (2018)

\bibitem{paszke2017automatic}
Paszke, A., Gross, S., Massa, F., Lerer, A., Bradbury, J., Chanan, G., Killeen,
  T., Lin, Z., Gimelshein, N., Antiga, L., Desmaison, A., Kopf, A., Yang, E.,
  DeVito, Z., Raison, M., Tejani, A., Chilamkurthy, S., Steiner, B., Fang, L.,
  Bai, J., Chintala, S.: {PyTorch}: An imperative style, high-performance deep
  learning library. In: Wallach, H., Larochelle, H., Beygelzimer, A.,
  d\textquotesingle Alch\'{e}-Buc, F., Fox, E., Garnett, R. (eds.) Adv. Neural
  Inform. Process. Syst., pp. 8024--8035. Curran Associates, Inc. (2019)

\bibitem{ren2015faster}
Ren, S., He, K., Girshick, R., Sun, J.: Faster {R-CNN}: Towards real-time
  object detection with region proposal networks. In: Adv. Neural Inform.
  Process. Syst. (2015)

\bibitem{szegedy2014intriguing}
Szegedy, C., Zaremba, W., Sutskever, I., Bruna, J., Erhan, D., Goodfellow, I.,
  Fergus, R.: Intriguing properties of neural networks. In: Int. Conf. Learn.
  Represent. (2014)

\bibitem{tashiro2020odi}
Tashiro, Y., Song, Y., Ermon, S.: Output diversified initialization for
  adversarial attacks. arXiv preprint arXiv:2003.06878  (2020)

\bibitem{tian2019fcos}
Tian, Z., Shen, C., Chen, H., He, T.: {FCOS}: Fully convolutional one-stage
  object detection. In: Int. Conf. Comput. Vis. (2019)

\bibitem{uesato2018adversarial}
Uesato, J., Odonoghue, B., Kohli, P., Den~Oord, A.V.: Adversarial risk and the
  dangers of evaluating against weak attacks. In: Int. Conf. Mach. Learn.
  (2018)

\bibitem{ulyanov2016in}
Ulyanov, D., Vedaldi, A., Lempitsky, V.: Instance normalization: The missing
  ingredient for fast stylization. arXiv preprint arXiv:1607.08022  (2016)

\bibitem{wang2020highfrequency}
Wang, H., Wu, X., Huang, Z., Xing, E.P.: High-frequency component helps explain
  the generalization of convolutional neural networks. In: IEEE Conf. Comput.
  Vis. Pattern Recog. (2020)

\bibitem{wang2004image}
Wang, Z., Bovik, A.C., Sheikh, H.R., Simoncelli, E.P.: Image quality
  assessment: From error visibility to structural similarity. IEEE Trans. Image
  Process.  \textbf{13}(4),  600--612 (2004)

\bibitem{wong2020ffgsm}
Wong, E., Rice, L., Kolter, J.Z.: Fast is better than free: Revisiting
  adversarial training. In: Int. Conf. Learn. Represent. (2020)

\bibitem{woo2018cbam}
Woo, S., Park, J., Lee, J.Y., So~Kweon, I.: {CBAM}: Convolutional block
  attention module. In: Eur. Conf. Comput. Vis. (2018)

\bibitem{wu2020awp}
Wu, D., Xia, S.T., Wang, Y.: Adversarial weight perturbation helps robust
  generalization. In: Adv. Neural Inform. Process. Syst. (2020)

\bibitem{wu2018gn}
Wu, Y., He, K.: Group normalization. In: Eur. Conf. Comput. Vis. (2018)

\bibitem{xu2020dctnet}
Xu, K., Qin, M., Sun, F., Wang, Y., Chen, Y., Ren, F.: Learning in the
  frequency domain. In: IEEE Conf. Comput. Vis. Pattern Recog. (2020)

\bibitem{zhang2019feature_scatter}
Zhang, H., Wang, J.: Defense against adversarial attacks using feature
  scattering-based adversarial training. In: Adv. Neural Inform. Process. Syst.
  (2019)

\bibitem{zhang2019theoretically}
Zhang, H., Yu, Y., Jiao, J., Xing, E.P., Ghaoui, L.E., Jordan, M.I.:
  Theoretically principled trade-off between robustness and accuracy. In: Int.
  Conf. Mach. Learn. (2019)

\bibitem{zhang2017beyond}
Zhang, K., Zuo, W., Chen, Y., Meng, D., Zhang, L.: Beyond a {Gaussian}
  denoiser: Residual learning of deep {CNN} for image denoising. IEEE Trans.
  Image Process.  \textbf{26}(7),  3142--3155 (2017)

\bibitem{zhou2018places}
Zhou, B., Lapedriza, A., Khosla, A., Oliva, A., Torralba, A.: Places: A 10
  million image database for scene recognition. IEEE Trans. Pattern Anal. Mach.
  Intell.  \textbf{40}(6),  1452--1464 (2018)

\end{thebibliography}
}
\end{document}